\documentclass{article}

\usepackage{graphicx} % more modern
\usepackage{subfigure}

\usepackage{natbib}

\usepackage{algorithm}
\usepackage{algorithmic}

\usepackage{amsmath}
\usepackage{amssymb}

\usepackage{hyperref}

\usepackage[accepted]{icml2013}
\icmltitlerunning{Hyperparameter Optimization and Boosting for Classifying Expressions}

\newcommand{\argmin}{\operatorname{argmin}}

\begin{document}

\twocolumn[
\icmltitle{Hyperparameter Optimization and Boosting for Classifying \\
           Facial Expressions: How good can a ``Null'' Model be?}

% unless you've provided the [accepted] option to the icml2013
\icmlauthor{James Bergstra}{bergstra@uwaterloo.ca}
\icmladdress{University of Waterloo,
            200 University Ave., Wateroo, Ont N2L 3G1 CANADA}
\icmlauthor{David D. Cox}{davidcox@fas.harvard.edu}
\icmladdress{Harvard University,
            52 Oxford St., Cambridge, MA 02138 USA}

\icmlkeywords{Computer Vision, Facial Expression Classification, Bayesian Optimization, Boosting}

\vskip 0.3in
]

%Another alternative is to use the \textbf{pdflatex} program instead of
%straight \LaTeX. This program avoids the Type-3 font problem, however
%you must ensure that all of the fonts are embedded (use {\tt
%pdffonts}). If they are not, you need to configure pdflatex to use a
%font map file that specifies that the fonts be embedded. Also you
%should ensure that images are not downsampled or otherwise compressed
%in a lossy way.

\begin{abstract}
One of the goals of the ICML workshop on representation and learning is to establish benchmark scores for a new data set of labeled facial expressions.
This paper presents the performance of a ``Null model'' consisting of convolutions with random weights, PCA, pooling, normalization, and a linear readout.
Our approach focused on hyperparameter optimization rather than novel model components.
On the Facial Expression Recognition Challenge held by the Kaggle website, our hyperparameter optimization approach achieved a score of 60\% accuracy on the test data.
This paper also introduces a new ensemble construction variant that combines hyperparameter optimization with the construction of ensembles.
This algorithm constructed an ensemble of four models that scored 65.5\% accuracy.
These scores rank 12th and 5th respectively among the 56 challenge participants.
It is worth noting that our approach was developed prior to the release of the data set, and applied without modification;
our strong competition performance suggests that the  TPE hyperparameter optimization algorithm and domain expertise encoded in our Null model
can generalize to new image classification data sets.
\end{abstract}

\section{Introduction}
\label{submission}

%These strong performances, together with the simplicity of the elements of our model class,
%indicate that the importance of hyperparameter optimization is widely underestimated.

The design of an effective machine learning system typically involves making many design choices that reflect the nature of data at hand and the inferences we wish to make.
The techniques at our disposal, as designers of machine learning systems, are
{\em intuition}: repeating what worked in other settings that seem to be similar (appealing to our expertise and intuition), and
{\em search}: model selection by trial and error search using e.g. cross-validation.

In common practice, both intuition and search are typically carried out informally.
A practitioner may design a complete system by a semi-automated search process in which
small-scale searches (e.g. grid search) update the practitioner's own implicit beliefs regarding
what constitutes a good model for the task at hand.
Those beliefs inform the choice of future small-scale searches in an iterative process that makes progressive improvements to the system.
(We may see many empirical results published in machine learning conferences as evidence of this process unfolding on an international scale over a course of years.)

One practical problem that arises from this common practice is that algorithms which have been demonstrated to work on particular data sets are notoriously difficult to adapt to new data sets.
The trouble is that the implicit beliefs of the practitioner play a crucial role in the process of model selection.
This difficulty has been widely recognized by domain experts, but the status quo remains because so many experts feel that their search is sufficiently efficient,
and the insights gained from the model selection process are valuable.
We hope that our results, taken together with other recent work on hyperparameter optimization such as \citet{bergstra+bardenet+bengio+kegl:2011,snoek+larochelle+adams:2012nips,thornton+hutter+hoos+leyton-brown:2012}, challenge these beliefs and induce more researchers to recognize automatic hyperparameter optimization as an important technique for model evaluation.

\subsection{Hyperparameter Optimization and Ensemble Construction}

While hyperparameter optimization is important, it is not the only standard performance-enhancing technique used to improve the scores of a model family on a given benchmark task.
Ensemble methods such as
Bagging \cite{breiman:1996bagging},
Boosting \cite{demiriz+bennett+shawe-taylor:2002},
Stacking \cite{wolpert:1992,breiman:1996stacked,sill+takacs+mackey+lin:2009},
and Bayesian Model Averaging \cite{hoeting+madigan+raftery+volinsky:1999}
are also commonly employed to exact every last drop of accuracy from a given set of algorithmic technology.
One of the goals of automating hyperparameter optimization is to assess, in an objective way, how good a set of classification system components can be.
In pursuit of that goal, the automation of ensemble creation is also critical.

This paper presents a first attempt to provide a fully automated algorithm for model selection and ensemble construction.
It starts from a palette of configurable pre-processing strategies, and classification algorithms, and proceeds to creates the most accurate ensemble of optimized components that it can.
Our algorithm uses a Boosting approach to ensemble construction, in which hyper-parameter optimization plays the role of a base learner.
This setting creates unique challenges that motivate a new Boosting algorithm, which we call SVM HyperBoost.

\section{Null Model for Image Classification}

Our basic approach is described in \citet{bergstra+yamins+cox:2013}.
We use Hyperopt \cite{hyperopt} to describe a configuration space that includes one-layer, two-layer, and three-layer convolutional networks.
The elements of our image classification model are standard scaling (image resolution), affine warp (rigid image deformation), filterbank normalized cross-correlation, local spatial pooling,
di-histogram spatial pooling, and an L2-SVM classifier.
For each layer in each architecture, hyperparameters govern the size of filters, the volume of pooling regions, constants that modulate local normalization, and so on.
The filters themselves are either chosen randomly from a centered Gaussian distribution, or are random projections of PCA components of training data (as it appears as input to each layer), 
or are random projections of input patches (again, as input arrives to each layer).
Features for classification are derived from the output layer by either signed (di-histogram) or unsigned pooling over some topographically local partitioning of output features.
This configuration space was chosen to span the model space investigated by \citet{pinto+cox:2011} and the random-filter models of \citet{coates+ng:2011}.
Relative to \citet{bergstra+yamins+cox:2013} we add the possibility of affine warping of input images and remove the input-cropping step.
In total, the configuration space includes 238 hyperparameters, although no configuration uses all 238 at once.
Many of the hyperparameters are {\em conditional} hyperparameters because they are only active in certain conditions;
for example, the hyperparameters governing the creation of a third layer are inactive for two-layer models.
%The largest number of hyperparameters that could be simultaneously active is
%around 60,
%which corresponds to a three-layer model with PCA-based filters at each layer, and one-sided ({\em di-histogram}) pooling at the output.
The details of this meta-model are described in \citet{bergstra+yamins+cox:2013} and implemented in the \texttt{hyperopt-convnet} software
available from \url{http://github.com/jaberg/hyperopt-convnet}.

Notable omissions from the model space include:
backpropagation \cite{Rumelhart86c},
unsupervised learning for filters such as
RBMs \cite{hinton:2002,hinton+osindero+teh:2006},
Sparse Coding \cite{coates+ng:2011},
DAAs \cite{vincent+larochelle+bengio+manzagol:2008},
and recent high-performance regularization strategies such as
dropout \cite{hinton+srivastava+krizhevsky+sutskever+salakhutdinov:2012}
and maxout \cite{goodfellow+warde-farley+mirza+courville+bengio:2013}.

Hyperparameter optimization within this Null model was carried out using the TPE algorithm \cite{bergstra+bengio:2012}, as implemented by
the \texttt{hyperopt} software (available from \url{http://jaberg.github.com/hyperopt}).

\section{SVM HyperBoosting}

This section describes an ensemble construction method (SVM HyperBoost, or just HyperBoost) that is particularly well-suited to the use of a hyperparameter optimization algorithm as an inner loop.
This algorithm is presented in the context of models which have the form of a feature extractor and a linear classifier.
In this context, the ensemble is simply a larger linear function that can be seen as the concatenation of ensemble members.
The HyperBoost algorithm can be understood as a piecewise training of this single giant linear classifier.

To derive the HyperBoost algorithm, suppose that we commit to using an ensemble of size $J$.
(No such commitment is necessary in practice, but it makes the development clearer.)
The ideal ensemble weights $\mathbf{w}^{(*)}$ and hyperparameter configuration settings $\lambda^{(*)}$
for a binary classification task would optimize generalization error:
\begin{align}
    \mathbf{w}^{(*)}, \mathbf{\lambda}^{(*)}  =&
    \argmin_{\mathbf{w} \in \mathbb{R}^*, \mathbf{\lambda} \in \mathbb{H}^J} ~\mathbb{E}_{\mathbf{x}, y \sim \mathcal{D}} 
    \nonumber \\
    & \left[ \mathbb{I} \{ 0 > y(\sum_j \mathbf{w}_j \cdot \mathbf{f}(x, \lambda_j) \} \right].
    \label{eq:svm_hyperopt}
\end{align}
Here $\mathcal{D}$ stands for a joint density over inputs and labels ($x \in \mathbb{R}^M$, $y \in \{-1, 1\}$).
We have used $\mathbf{f}(x, \lambda_j)$ to denote the feature vector associated to input $x$ by hyperparameter configuration $\lambda_j$.
The expression $\mathbb{I}\{ 0 > a \}$ denotes the indicator function that is one for values of $a$ which are negative, but zero for values of $a$ which are not negative.
We use $\mathbb{H}$ to stand for the set of possible hyperparameter configurations, so that the argmin means ``choose $J$ optimal hyperparameter configurations'' (one for each ensemble member).
We use the notation $\mathbf{w} \in \mathbb{R}^*$ in the argmin to indicate that the final set of weights $\mathbf{w}^{(*)}$ will be a vector, but it is not known a-priori how many elements it will have.
Rather, $\mathbf{w}$ will be logically divided into $J$ pieces corresponding to ensemble elements and each piece $\mathbf{w}_j$ will have a dimensionality that matches $\mathbf{f}(x; \lambda_j)$.

\begin{figure}[ht]
\vskip 0.2in
\begin{center}
\centerline{\includegraphics[width=.8\columnwidth]{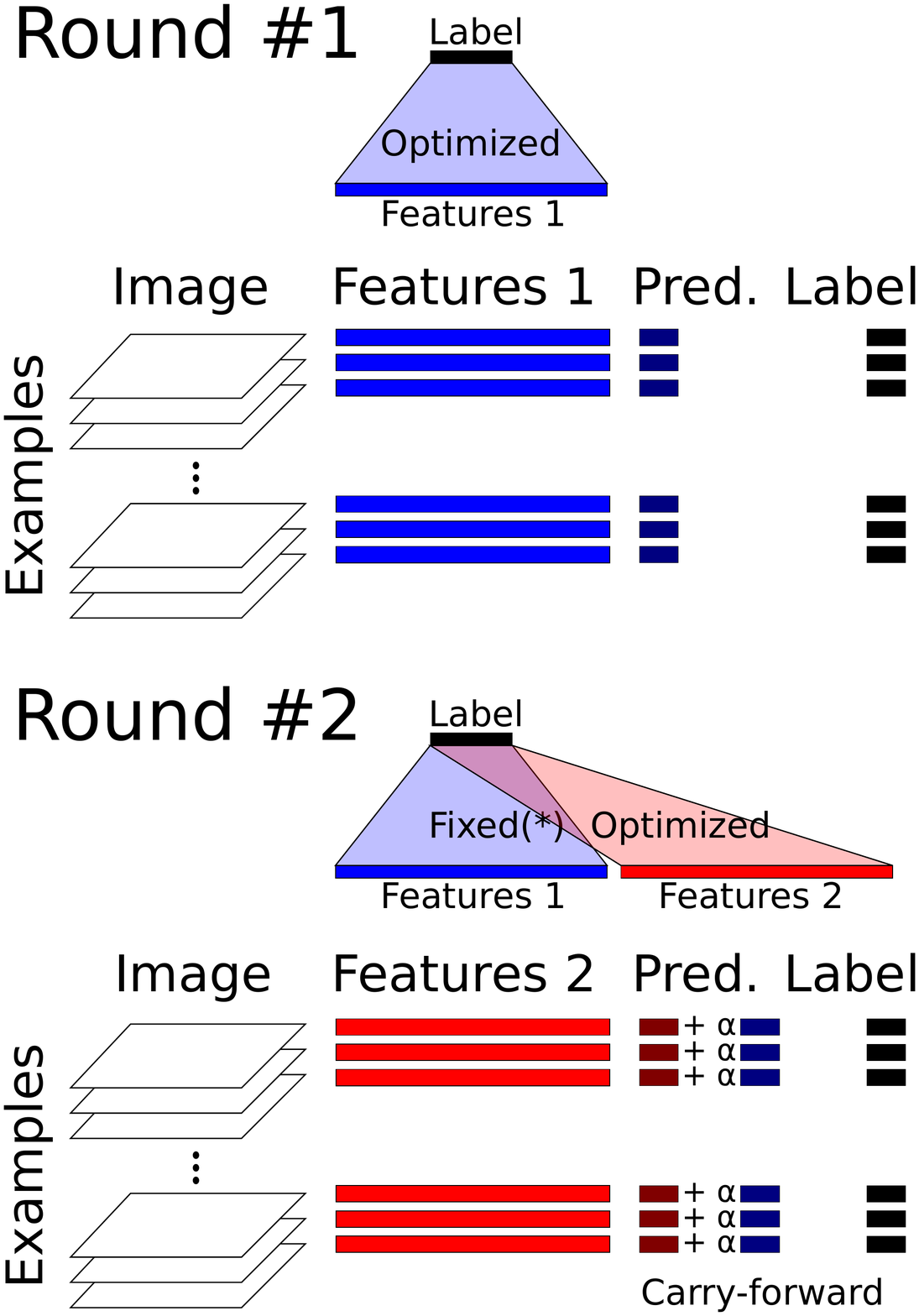}}
\caption{
    The SVM HyperBoost algorithm creates a large linear SVM piece-wise. The first round of training is standard SVM training.
    At the end of the first round, the SVM weights are fixed (*) up to multiplicative scaling.
    Subsequent rounds ``carry forward'' the total contribution of previous features and their corresponding fixed weights toward label predictions.
    On Round 2, HyperBoost optimizes the feature weights (shown in light red)
    for a candidate feature set (bright red) and re-scales (via $\alpha$) weights fit in previous rounds.
    This approximate, greedy procedure makes it possible to fit very large SVMs to large numbers of examples, when feature computation is also computationally costly.
   }
\label{fig:isvm}
\end{center}
\vskip -0.2in
\end{figure}

The joint optimization of $\mathbf{w}$ and $\lambda$ implied by Equation~\ref{eq:svm_hyperopt} is challenging because of
\begin{itemize}
    \item the complicated effect of each $\lambda_j$ on $\mathbf{f}$,
    \item the expectation over unknown $\mathcal{D}$, and
    \item the non-differentiable indicator function.
\end{itemize}
Our strategy for dealing with the complicated relationship between $\lambda_j$ and $\mathbf{f}$ is to select the $\lambda_j$ configurations greedily, using the algorithm illustrated in Figure~\ref{fig:isvm}.
Our strategy for dealing with the expectation is to estimate it from what is typically called validation data, so that each argmin for $j < J$ is what previous work has called hyperparameter optimization.
Our strategy for dealing with the non-differentiability of the indicator function is to use a gradient-free optimization method, namely TPE \cite{bergstra+bardenet+bengio+kegl:2011}.

Normally, Boosting (functional gradient methods) on a Hinge loss or Zero-One loss would quickly run into trouble because once the training margins are pushed past the decision boundary, subsequent rounds have nothing to do (the training criterion is completely satisfied).
We avoid this nonsense using two techniques.
First, Boosting on sufficient validation data helps because models fit to training data are seldom perfect for validation data by random chance
(incidentally, we are interested in collaborations that might shed light on exactly how much validation data is necessary).
Second, each round of HyperBoost is free to scale the contribution of previous ensemble components (see $\alpha$ in Figure~\ref{fig:isvm}), so standard SVM regularization techniques (i.e. $C$) allow us to meaningfully add features and improve $\mathbf{w}$ even if the Hinge loss had been reduced to 0 at a previous HyperBoosting iteration.
The regularization parameter governing the entire SVM is a hyperparameter that is re-optimized on every round of HyperBoost.
This technique makes HyperBoost a partially corrective Boosting algorithm.

\subsection{Weak Learners vs. Strong Learners}

HyperBoost is suitable for Boosting {\em strong} base learners.
In fact, when Boosting and model fitting are conducted on statistically independent example sets, the distinction between distinction between ``weak'' vs. ``strong'' learners is no longer important.
Instead, any learners (weak or strong) simply provide models, and HyperBoosting chooses the model that most improves the validation set performance of the ensemble.
While strong learners generally require additional regularization compared with weak learners in order to generalize correctly from training data,
strong and weak base learners are equally useful for HyperBoosting.

\section{Results on Facial Expression}

In support of the ICML2013 workshop on representation learning,
Pierre-Luc Carrier and Aaron Courville released a data set for facial expression recognition as a Kaggle competition (
\texttt{http://www.kaggle.com/c/challenges-in-\\
representation-learning-facial-expression-\\
recognition-challenge}).
The data consist of 48x48 pixel grayscale images of faces, and labels for the expressions of those faces.
The faces have been automatically registered so that each face is approximately centered and occupies about the same amount of area within each image.
The task is to categorize each face as one of seven categories (anger, disgust, fear, happy, sad, surprised, neutral).
The set distributed by Kaggle consists of 28,709 training examples examples, and 7,178 test examples.

Our protocol for model selection was simple cross-validation on the training examples.
We partitioned the training data into 20709 SVM-fitting examples and 8000 validation examples,
and performed hyperparameter optimization with regard to the performance on this validation set.
The test examples were not used for model selection.
The Kaggle website only provided the images for test examples, to prevent cheating in the contest.
The test scores listed for the Facial Expression Recognition Challenge were obtained by uploading our predictions to Kaggle's website, which computed the test set accuracy on our behalf.

On each round of HyperBoosting, we evaluated 1000 non-degenerate hyperparameter proposals in search of the best feature set to add to the ensemble.
These proposals ranged in accuracy from chance baseline of 20\% up to a relatively strong 62\%.
Experiments were done using a single computer with four NVidia Tesla 2050 GPUs and a slow file system so typically 2 or 3 jobs would run simultaneously.
Many configurations were invalid (e.g. downsampling so much that 0 features remain) but these are recognized relatively quickly.
Valid (non-degenerate) trials typically took 10 - 25 minutes to complete, so each round of HyperBoost took two or three days using this one machine.

\begin{figure}[ht]
\vskip 0.2in
\begin{center}
    \centerline{{\bf HyperBoost for Ensemble Construction}}
    \centerline{\includegraphics[width=\columnwidth]{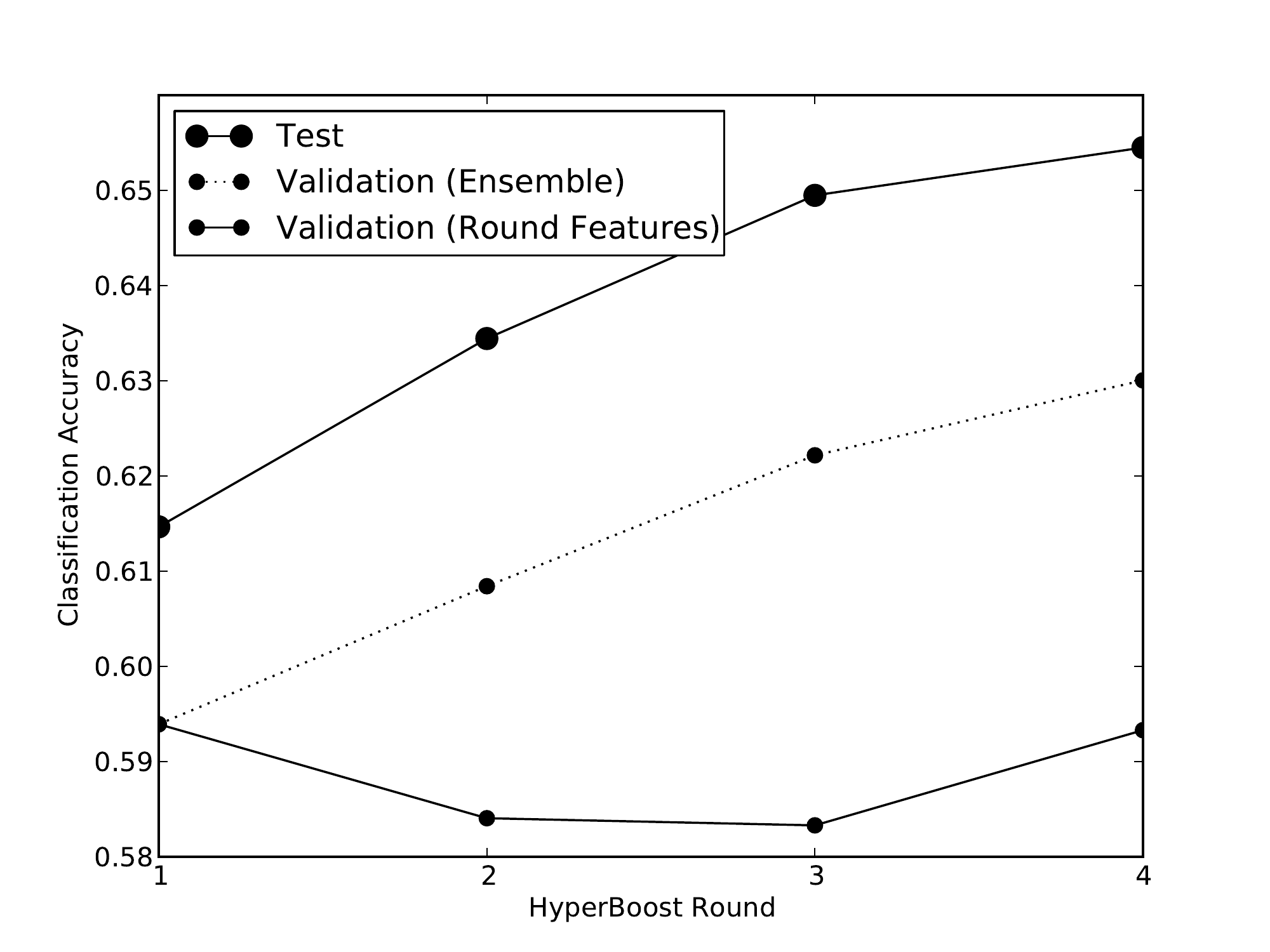}}
    \caption{HyperBoost improves test set generalization with successive rounds while the individual feature sets chosen at each round hold steady just below 60\% accuracy .
        Each round selects the best of 1000 non-degenerate candidate feature sets.
        Training set accuracy (not shown) ranges from 85\% for the first round up to 97\% on the fourth round.
    }
\label{fig:hyperboost}
\end{center}
\vskip -0.2in
\end{figure}

\begin{table}[t]
\caption{Performance relative to Kaggle submissions.}
\label{tbl:rankings}
\vskip 0.15in
\begin{center}
\begin{small}
\begin{sc}
\begin{tabular}{llc}
\hline
\abovespace\belowspace
Rank & Team & Accuracy (\%) \\
\hline
\abovespace
1 & ``RBM''                & $71.162$ \\
2 & ``Unsupervised''       & $69.267$ \\
3 & ``Maxim Milakov''      & $68.821$ \\
4 & ``Radu+Marius+Cristi'' & $67.484$ \\
- & {\bf HyperBoost Round 4} & $65.450$ \\
5 & ``Lor.Voldy''          & $65.255$ \\
$\vdots$ \\
11 & ``jaberg''            & $61.967$ \\
 - & {\bf HyperBoost Round 1} & $61.466$ \\
12 & ``bulbugoglu''        & $59.654$ \\
$\vdots$ \\
\belowspace
56 & ``dstarerstor'' & $20.006$ \\
\hline
\end{tabular}
\end{sc}
\end{small}
\end{center}
\vskip -0.1in
\end{table}

The accuracies of the models chosen by HyperBoost are shown in Figure~\ref{fig:hyperboost}.
HyperBoost creates a small ensemble whose combined accuracy (65.5\%) is significantly better than the best individual model (60\%).
The ranking relative to other models in the Kaggle competition is shown in Table~\ref{tbl:rankings}.
The ensemble of size 4 ranks among the top 5 competition entries.

It is worth noting that the model and training programs used for HyperBoosting in this model space were entirely designed prior to the release of the data set.
The model space was chosen to span the models of \citet{pinto+cox:2011} and \citet{coates+ng:2011}.
\citet{pinto+cox:2011} reported excellent match verification performance on the Labeled Faces in the Wild (LFW) data set \cite{huang+ramesh+berg+learnedmiller:2007}),
and \citet{coates+ng:2011} advanced the state of the art at the time on the CIFAR-10 object recognition data set.
Our approach was developed prior to the release of the Facial Expression Recognition data set,
so the good performance speaks directly to the ability of our meta-modeling approach to generalize to new image classification tasks.

It is also worth noting that the training accuracy (not shown) of all models in the ensemble is much higher than the generalization accuracy (from $85\%$ up to $97\%$).
The size of individual feature sets was capped at 9000 (to stay within the available memory on the GPU cards), and all of the best models approached this maximum number of features.
Although these large feature sets demonstrated significant over-fitting of the training data (these feature sets represent {\em strong} base learners for Boosting),
HyperBoost selected ensemble members that brought steady improvement on the test set.
This is a familiar story for Boosting algorithms based on an exponential loss, but HyperBoost produces the effect while operating on the more representative hinge loss.

\section{Conclusion}

Hyperparameter optimization within large model classes is difficult.
We have shown that hyperparameter optimization within a Null model achieves over 60\% accuracy in the workshop's Facial Expression Recognition Challenge,
which ranks 12th of 56 contest submissions.
Further use of an ensemble-construction mechanism raises that accuracy to 65.5\%, which would have ranked 5th / 56 had it been ready by the contest closing date.
These performances underscore the importance and difficulty of fully leveraging known algorithmic technology for image classification.
We can only conjecture that a future version of our model space that includes a wider range of algorithms for feature initialization and refinement (e.g.
backpropagation, dropout, maxout, sparse coding, sparsity regularization, RBMs, DAAs)
could perform better yet.
By the same token, until such a search is carried out, it is difficult to make quantitative claims regarding the value added by such algorithms over and above a well-configured set of simpler components.

The software used in these experiments is publicly available from github:
\begin{description}
    \item[Hyperopt] The TPE hyperparameter optimization algorithm and distributed optimization infrastructure.  \url{http://jaberg.github.com/hyperopt}
    \item[Hyperopt-ConvNet] The HyperBoost algorithm and hyperopt-searchable representation of the image classification model. \url{http://github.com/jaberg/hyperopt-convnet}
\end{description}

\section*{Acknowledgments}

This project has been supported by the Rowland Institute of Harvard, the United States' National Science Foundation (IIS 0963668), and
Canada's National Science and Engineering Research Council through the Banting Fellowship program.

\bibliography{example_paper,../bmvc/paper}
\bibliographystyle{icml2013}

\end{document}